\documentclass{article}
\usepackage{spconf,amsmath,graphicx}
\usepackage{booktabs,multirow}
\usepackage{amssymb}
\usepackage{hyperref}
\usepackage{cleveref}
\usepackage{siunitx} 
\usepackage{tabularx}

\usepackage{multirow}

\usepackage{enumitem}
\usepackage{bm}
\usepackage{color}
\usepackage{makecell}
\usepackage{setspace}

\usepackage[table]{xcolor}

\let\OLDthebibliography\thebibliography
\renewcommand\thebibliography[1]{
  \OLDthebibliography{#1}
  \setlength{\parskip}{0pt}
  \setlength{\itemsep}{0pt plus 0.3ex}
}

\begin{document}\sloppy

\def\x{{\mathbf x}}
\def\L{{\cal L}}

\title{DARA: Domain- and Relation-aware Adapters Make Parameter-efficient Tuning for Visual Grounding}
%

\name{\begin{tabular}{c}{}
Ting Liu$^{1*}$
Xuyang Liu$^{2*}$
Siteng Huang$^3$
Honggang Chen$^2$
Quanjun Yin$^1$ \\
Long Qin$^1$
Donglin Wang$^3$
Yue Hu$^{1\dag}$
\thanks{$^*$Equal contribution. $^\dag$Corresponding author.}
\end{tabular}}

\address{$^1$College of Systems Engineering, National University of Defense Technology, Changsha, China\\
$^2$College of Electronics and Information Engineering, Sichuan University, Chengdu, China\\
$^3$School of Engineering, Westlake University, Hangzhou, China \\
\tt\small liuting20@nudt.edu.cn, liuxuyang@stu.scu.edu.cn}

\maketitle

\begin{abstract}
Visual grounding (VG) is a challenging task to localize an object in an image based on a textual description. Recent surge in the scale of VG models has substantially improved performance, but also introduced a significant burden on computational costs during fine-tuning. In this paper, we explore applying parameter-efficient transfer learning (PETL) to efficiently transfer the pre-trained vision-language knowledge to VG. Specifically, we propose \textbf{DARA}, a novel PETL method comprising \underline{\textbf{D}}omain-aware \underline{\textbf{A}}dapters (DA Adapters) and \underline{\textbf{R}}elation-aware \underline{\textbf{A}}dapters (RA Adapters) for VG. DA Adapters first transfer intra-modality representations to be more fine-grained for the VG domain. Then RA Adapters share weights to bridge the relation between two modalities, improving spatial reasoning. Empirical results on widely-used benchmarks demonstrate that DARA achieves the best accuracy while saving numerous updated parameters compared to the full fine-tuning and other PETL methods. Notably, with only \textbf{2.13\%} tunable backbone parameters, DARA improves average accuracy by \textbf{0.81\%} across the three benchmarks compared to the baseline model. Our code is available at \url{https://github.com/liuting20/DARA}.
\end{abstract}
\begin{keywords}
Visual grounding, parameter-efficient transfer learning, vision-language models
\end{keywords}
\section{Introduction}
\label{sec:intro}

Visual grounding (VG) \cite{yu2016modeling,mao2016generation} is one of the most challenging tasks in the vision-language learning fields. Compared with tasks like text-image retrieval and image captioning, VG requires more fine-grained vision-language alignment for spatial reasoning to accurately locate an object described by a language expression. Existing VG methods \cite{yu2018mattnet,yang2019fast,deng2021transvg,zhang2023one} follow a standard \textit{pretrain-then-finetune} paradigm, which first pre-trains the vision and language backbones on large-scale datasets and then fine-tunes the whole models on downstream specific datasets \cite{yu2016modeling,mao2016generation}. As the backbone networks of VG models become increasingly deeper, the number of parameters grows rapidly, from 47M in MAttNet \cite{yu2018mattnet} to 151M in TransVG \cite{deng2021transvg}. Fine-tuning such large VG models can be computationally expensive and time-consuming in practice. 

\begin{figure}[t]
\centering
\includegraphics[width=1\linewidth]{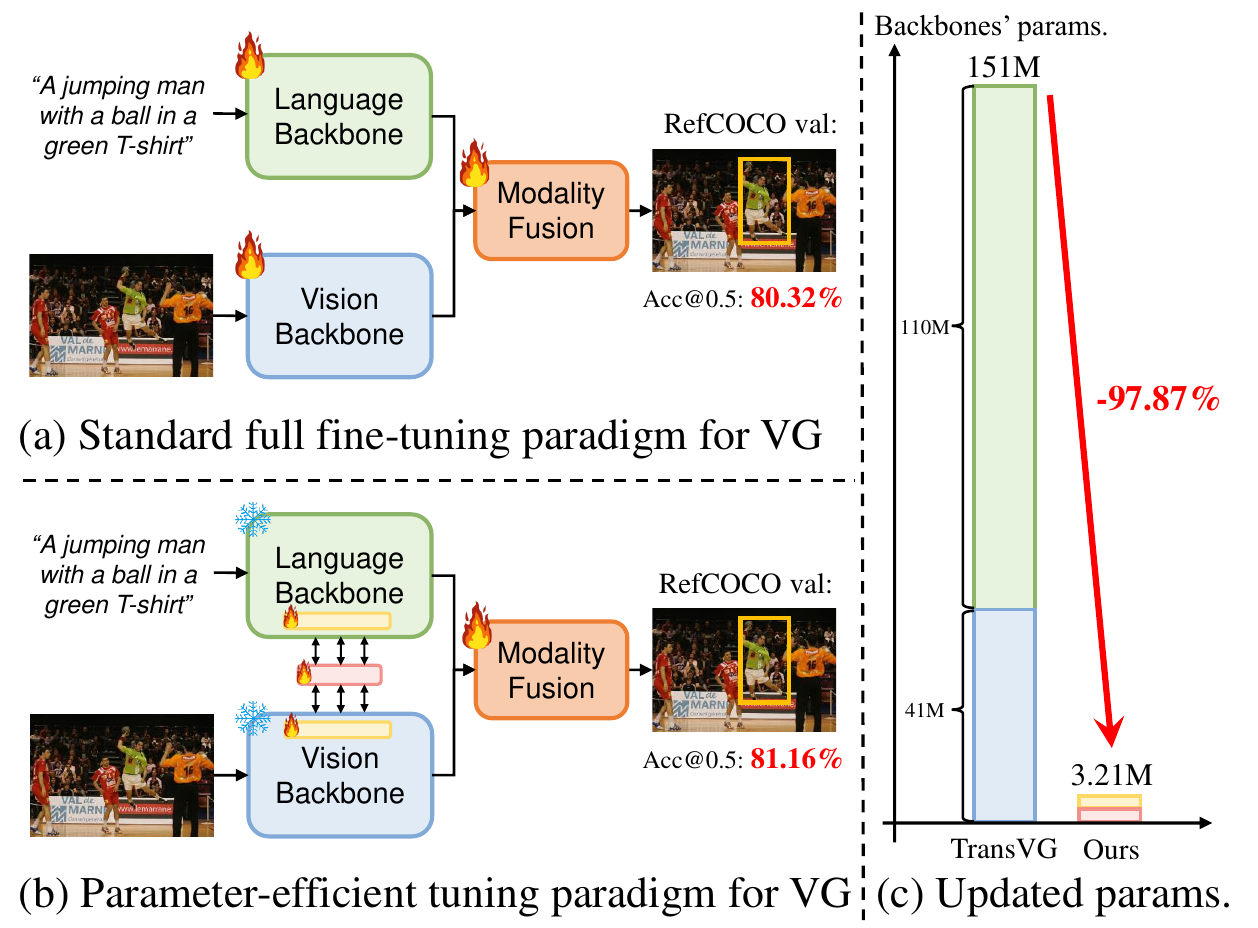}
\vspace{-7mm}
\caption{Comparison of (a) full fine-tuning \cite{deng2021transvg} and (b) our PETL method for visual grounding. (c) Freezing (\protect\includegraphics[height=0.3cm]{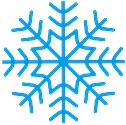}) the pre-trained backbones and updating (\protect\includegraphics[height=0.3cm]{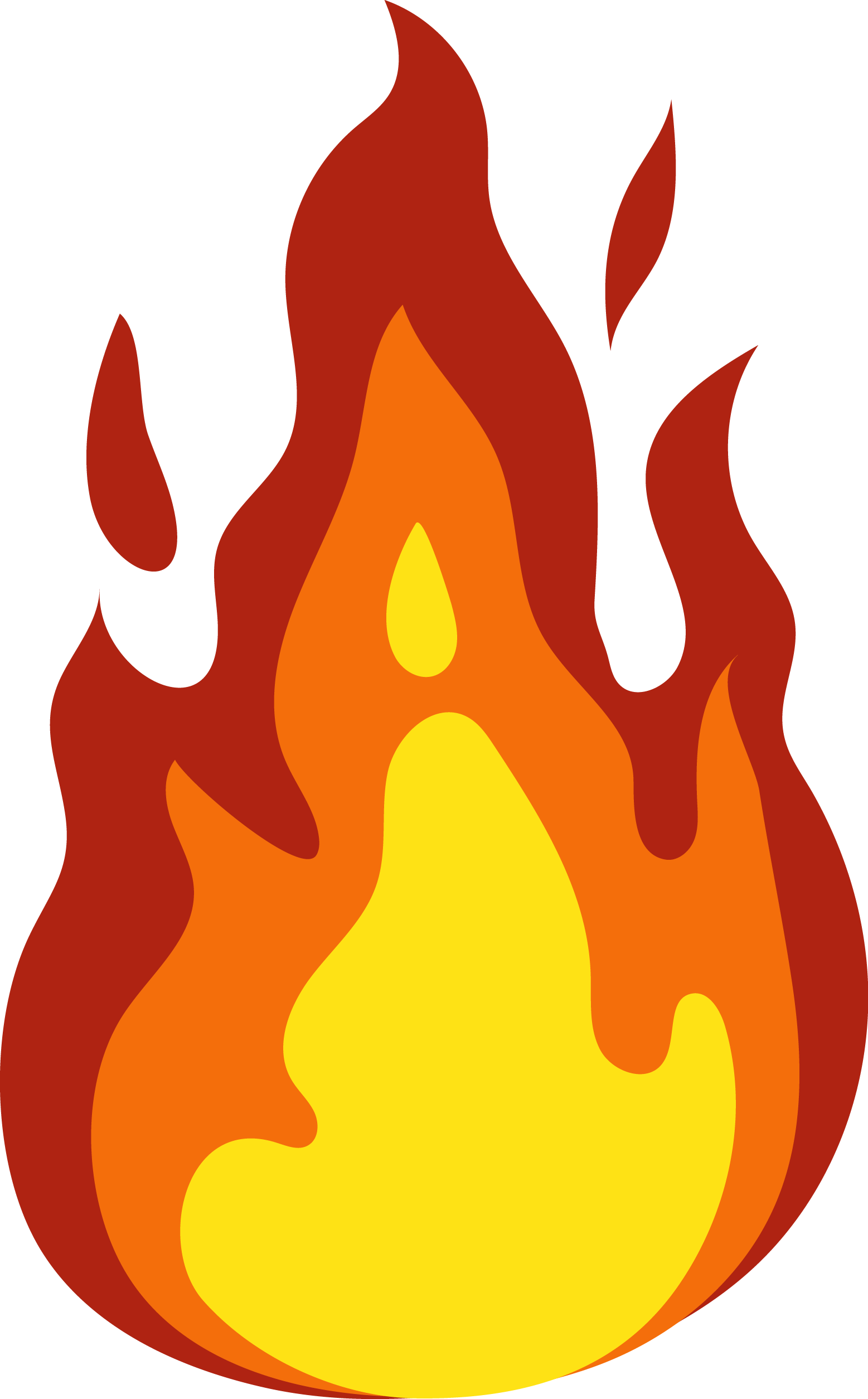}) our DARA reduces ~\textbf{97.87\%} of backbone updated parameters while achieving even \textbf{stronger} performance than full fine-tuning paradigm.}
\vspace{-3mm}
\label{fig1}
\end{figure}

Recently, a variety of parameter-efficient transfer learning (PETL) methods \cite{houlsby2019parameter,chen2022adaptformer,hu2022yelong} have demonstrated high efficiency and extensibility in transferring diverse knowledge from pre-training to downstream tasks. Specifically, these PETL methods freeze nearly all parameters of the pre-trained model, and only update a small number of additional adaptable parameters. With just a tiny fraction of trainable parameters, they can achieve comparable performance to full fine-tuning. Current PETL methods are proposed to address single-modality tasks or simple cross-modal tasks, such as language understanding \cite{houlsby2019parameter}, image classification \cite{chen2022adaptformer} and text-image/video retrieval \cite{yuan2023parameter,jiang2022cross,huang2023vop}. This gives rise to an intriguing question: \textit{can these PETL methods be directly applied to the visual grounding task to reduce the computational and time cost of fine-tuning?}. Experiments in \Cref{sec:Main Results} demonstrate that directly applying them to the VG task can lead to considerable performance degradation. We suppose that the PETL methods lack sufficient interaction between the visual and linguistic representations for spatial reasoning.

Motivated by the above analysis, this paper aims to effectively and efficiently fine-tunes a standard transformer-based VG model to match the full fine-tuning results but with remarkably decreased computational cost. Specifically, we propose a novel adapter-based PETL method termed \textbf{DARA} that incorporates \underline{\textbf{D}}omain-aware \underline{\textbf{A}}dapters (DA Adapters) and \underline{\textbf{R}}elation-aware \underline{\textbf{A}}dapters (RA Adapters), for visual grounding. DA Adapters first transfer the pre-trained intra-modality representations in the dual backbones to more fine-grained vision and language representations to the VG \textbf{domain}. Subsequently, RA Adapters enable early cross-modal interactions by sharing the adapters' weights between the two backbones, thus bridging the \textbf{relation} between the two modalities which facilitates the cross-model reasoning abilities of the baseline model. DARA based on a pre-trained VG method, TransVG \cite{deng2021transvg}, can reduce approximately \textbf{97.87\%} updated parameters of the pre-trained vision and language backbones while still achieving comparable performance with the full fine-tuning paradigm, as shown in \Cref{fig1}. Furthermore, we surprisingly discover that the early cross-modal interactions enabled by our proposed RA Adapters can improve fusion effectiveness for spatial reasoning, which is validated by the experiments depicted in \Cref{fig3}.

Our main contributions can be summarized as threefold: 
\begin{itemize}[noitemsep,nolistsep]
	\item We present an in-depth exploration of adapting parameter-efficient transfer learning (PETL) methods for visual grounding. To the best of our knowledge, this is the first study to investigate parameter-efficient tuning paradigm for visual grounding.
	\item We propose DARA, a novel PETL framework incorporating two tailored adapters: Domain-aware Adapters and Relation-aware Adapters, to facilitate effective and efficient intra- and inter-modality representation transfer for the visual grounding task.
	\item Extensive experiments on three widely-used VG benchmarks demonstrate that DARA can achieve the best accuracy while saving numerous updated parameters compared to full fine-tuning and other PETL methods.
\end{itemize}

\section{Related Work}
\label{sec:Related Work}

\noindent \textbf{Visual Grounding.} Visual grounding (VG) \cite{yu2016modeling,mao2016generation} is a challenging vision-language modeling task that aims to localize the specific object in an image by the given textual description. Currently, most VG methods can be grouped into three types based on their network architectures: (1) \textbf{Two-stage} VG methods \cite{yu2018mattnet,hong2019learning,liu2024vgdiffzero} first employ a pre-trained object detector to generate series of sparse region proposals, and then select the most similar one to the description for grounding. (2) \textbf{One-stage anchor-based} VG methods \cite{yang2019fast,ye2021one,sun2022proposal} integrate detect-then-select progress into one single pipeline that jointly fuses visual and textual representations at the intermediate layers of the detector and predicts the box by the pre-defined anchors. (3) \textbf{Transformer-based} VG methods \cite{deng2021transvg,du2022visual,zhang2023one} are proposed for end-to-end VG, benefiting from their strong intra- and inter-modality representation learning. Unfortunately, fine-tuning recent VG methods is becoming far more computationally expensive and time-consuming.

\noindent \textbf{Parameter-efficient Transfer Learning.} Parameter-efficient transfer learning (PETL) \cite{houlsby2019parameter,hu2022yelong} has become a prevalent approach for efficiently transferring fixed pre-trained models to downstream tasks by fine-tuning just a few parameters. PETL methods can be broadly categorized into two types: (1) Updating additional parameters in the module inserted into the model \cite{houlsby2019parameter,chen2022adaptformer} or appended with the input data \cite{liu2023dap,xin2024mmap}; (2) Decomposing the weight matrices of pre-trained models into two low-rank matrices and only updating the small factorization matrices \cite{hu2022yelong}. Recently, adapter-based PETL methods have shown strong efficiency, extensibility, and performance in various downstream vision-language tasks \cite{jiang2022cross,yuan2023parameter,xu2023bridging}. Our work focus on efficiently fine-tuning a pre-trained transformer-based VG method by carefully designed two types of adapters, which facilitate efficient intra- and inter-modality representation transferring.

\section{Methodology}
\label{sec:Methodology}

\subsection{Baseline Model}
\label{sec:Baseline Model}

\begin{figure*}[h]
\centering

\includegraphics[width=\textwidth]{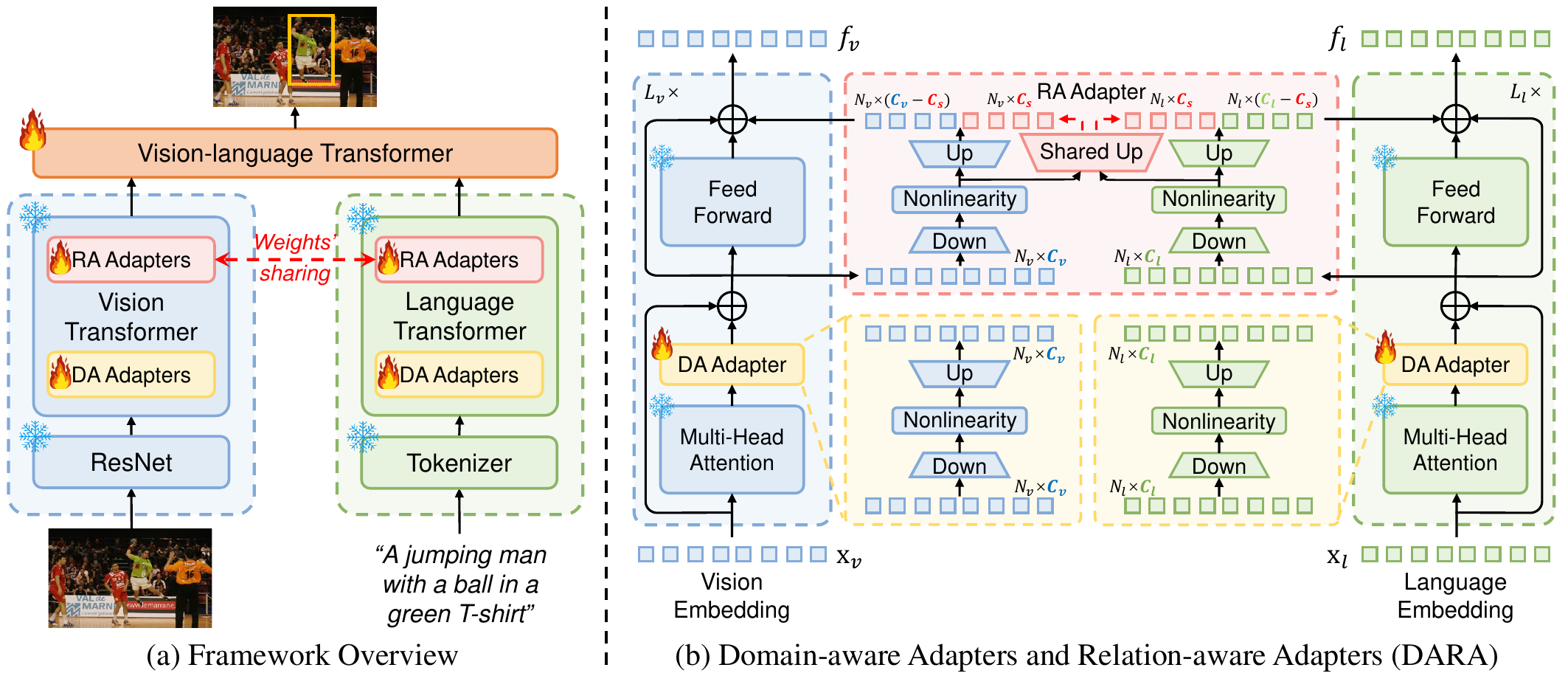}
\vspace{-7mm}
\caption{(a) Overview of our proposed parameter-efficient tuning framework for visual grounding. We freeze (\protect\includegraphics[height=0.3cm]{ice.png}) the vision and language backbones and update (\protect\includegraphics[height=0.3cm]{fire.png}) our DARA, comprising the Domain-aware Adapters (DA Adapters) and Relation-aware Adapters (RA Adapters). (b) Detailed design of the DARA. The DA Adapters transfer the pre-trained rich intra-modality representations, making them more fine-grained for the visual grounding domain. Subsequently, the RA Adapters then share the adapters' weights to bridge the relation between the two backbones and capture inter-modality representations.}
\vspace{-3mm}
\label{fig2}
\end{figure*}

As depicted in \Cref{fig2} (a), we follow the architecture and training objective of TransVG \cite{deng2021transvg}, a typical transformer-based VG method, to implement our model. It consists of three components: (1) Vision Backbone, (2) Language Backbone, and (3) Vision-language Transformer.  

\noindent \textbf{Vision Backbone.} The Vision Backbone, comprising a pre-trained ResNet \cite{he2016deep} and a vision transformer, encodes the image into a high-quality vision embedding. Specifically, given an input image $\bm{z}_0 \in \mathbb{R}^{H_0 \times W_0 \times 3}$, the Vision Backbone first uses the ResNet to generate a 2D feature map $\bm{z} \in \mathbb{R}^{H \times W \times C}$, where $C=2048$ is the channel dimension, $H=\frac{H_0}{32}$, and $W=\frac{W_0}{32}$. A $1\times 1$ convolutional layer then reduces $C$ to $C_v=256$, producing $\bm{z}' \in \mathbb{R}^{H \times W \times C_v}$. After flattening $\bm{z}'$ into the vision embedding $\bm{z}_v \in \mathbb{R}^{N_v \times C_v}$, where $N_v=H\times W$ represents the number of input tokens, $\bm{z}_v$ added with the position embedding is input to the vision transformer including 6 stacked transformer encoder layers. The vision transformer performs parallel processing on $\bm{z}_v$ and outputs an enhanced vision embedding $\bm{f}_v \in \mathbb{R}^{N_v \times C_v}$.

\noindent \textbf{Language Backbone.} The Language Backbone, adopting a pre-trained BERT \cite{devlin2018bert}, encodes the language expression into a language embedding. Specifically, given a language expression, the Language Backbone first converts each word ID into a one-hot vector, then tokenizes these vectors into series of language tokens. The language tokens, concatenated with a \texttt{[CLS]} token at the beginning and a \texttt{[SEP]} token at the end, are input to the language transformer comprising 12 identical transformer encoder layers. Similar to the vision transformer, the language transformer outputs an enhanced language embedding $\bm{f}_l \in \mathbb{R}^{N_l \times C_l}$, where $N_l$ and $C_l$ represent the number and channel dimension of language tokens, respectively.

\noindent \textbf{Vision-language Transformer.} The Vision-Language Transformer (V-L Transformer) aims to fuse the two modality embeddings and output a \texttt{[REG]} token for generating the bounding box. Specifically, after the vision and language backbones output the enhanced vision embedding $\bm{f}_v \in \mathbb{R}^{N_v \times C_v}$ and language embedding $\bm{f}_l \in \mathbb{R}^{N_l \times C_l}$, these embeddings are first projected into the same channel dimension $C_p=256$. The joint embeddings $\bm{f'}_v \in \mathbb{R}^{N_v \times C_p}$ and $\bm{f'}_l \in \mathbb{R}^{N_l \times C_p}$ are then concatenated with a learnable \texttt{[REG]} token, forming the input $\bm{x}_0 \in \mathbb{R}^{(N_v+N_l+1) \times C_p}$ for the V-L Transformer with a stack of 6 transformer encoder layers. Sequentially, the V-L Transformer outputs the \texttt{[REG]} token to a Prediction Head, implemented as a MLP with two 256-dimensional hidden layers with ReLU activation and a linear output layer, to regress the 4-dimensional box coordinates for the referred object.

\subsection{Domain-aware and Relation-aware Adapters}
\label{sec:DARA}

As the vision and language backbones contain most model parameters (around 95\%) and have acquired rich knowledge during pre-training, we attempt to freeze them during fine-tuning. However, fully freezing the backbones deteriorates visual grounding performance, as shown in \Cref{Table:ablations}. We hypothesize that is because the pre-trained backbones lack \textbf{two key abilities} required for adapting to the visual grounding task: (1) Fine-grained intra-modality representations for precisely describing and locating the objects. (2) Early cross-modal interactions between the vision and language backbones to enable joint spatial reasoning. Based on the above analysis, we propose DARA with Domain-aware Adapters and Relation-aware Adapters for visual grounding, as shown in \Cref{fig2}. Note that we do not show the LayerNorm for simplicity.

\noindent \textbf{DA Adapters.} The core function of DA Adapters is to transfer the pre-trained rich and general intra-modality representations to more fine-grained representations for the VG domain.  Specifically, the architecture of the DA Adapters follows the design of Adapters \cite{houlsby2019parameter} in natural language processing (NLP), comprising a down-projection layer, a non-linear layer, and an up-projection layer. The DA Adapters are inserted after the Multi-Head Attention layers (MHA) in the vision and language transformers. Taking the DA Adapters in the vision transformer as an example, formally, given the visual feature map $\bm{x}_v \in \mathbb{R}^{N_v \times C_v}$ processed by the MHA, the function of DA Adapter in the transformer encoder can be expressed as:
\begin{equation}
\text{DA\ Adapter}(x_v) = x_v + s\cdot{\sigma(x_v\mathbf{W}_{\text{down}})}\mathbf{W}_{\text{up}},
\end{equation}
where $\mathbf{W}_{\text{down}} \in \mathbb{R}^{C_v \times C_d}$ and $\mathbf{W}_{\text{up}} \in \mathbb{R}^{C_d \times C_v}$ are the weights of down- and up-projection layers, $\sigma(\cdot)$ is the ReLU activation function, and $s$ is the scaling factor. In this way, DA Adapters can refine the rich pre-trained vision and language representations into more fine-grained intra-modality representations for the VG domain during fine-tuning.

\noindent \textbf{RA Adapters.} Inspired by the existing efforts \cite{jiang2022cross}, our RA Adapters are designed to leverage the inter-modality representations, enabling early interaction between the two separate backbones, as shown in \Cref{fig2} (b). Specifically, the RA Adapter is composed of a down-projection layer, a non-linear layer, a unique up-projection layer, and a shared up-projection layer. Unlike the CM Adapters \cite{jiang2022cross}, our RA Adapters are connected \textbf{in parallel} with the Feed Forward Networks (FFN) in both vision and language backbones to enhance the fine-tuning effectiveness, as evidenced in \Cref{Table:comparison of insertaion forms}. Once the DA Adapters in both transformer encoders generate the enhanced embeddings, these embeddings are added with the original ones through skip-connections. Formally, given these adapted vision embedding $\bm{x}_v \in \mathbb{R}^{N_v \times {C_v}}$ and language embedding $\bm{x}_l \in \mathbb{R}^{N_l \times {C_l}}$, the down-projection layers first down-sample them to obtain the bottleneck features $\bm{z}_v \in \mathbb{R}^{N_v \times C_d}$ and $\bm{z}_l \in \mathbb{R}^{N_l \times C_d}$ as follows:
\begin{equation}
z_v = \sigma(x_v\mathbf{W}_{\text{down,vision}}),
\end{equation}
\begin{equation}
z_l = \sigma(x_l\mathbf{W}_{\text{down,language}}),
\end{equation}
where $\mathbf{W}_{\text{down,vision}} \in \mathbb{R}^{{C_v} \times C_d}$, $\mathbf{W}_{\text{down,language}} \in \mathbb{R}^{{C_l} \times C_d}$, and $\sigma(\cdot)$ is the ReLU activation function. Then, the up-projection layers with the sharing mechanism up-sample these bottleneck features to obtain the inter-modality features $\bm{f}_v \in \mathbb{R}^{N_v \times {C_v}}$ and $\bm{f}_l \in \mathbb{R}^{N_l \times {C_l}}$ as follows:
\begin{equation}
f_v = \text{Concat}[z_v{\mathbf{W}_{\text{up,vision}}}, z_v{\mathbf{W}_{\text{up,share}}}],
\end{equation}
\begin{equation}
f_l = \text{Concat}[z_l{\mathbf{W}_{\text{up,language}}}, z_l{\mathbf{W}_{\text{up,share}}}],
\end{equation}
where ${\mathbf{W}_{\text{up,vision}}} \in \mathbb{R}^{C_d \times ({C_v}-{C_s})}$ and ${{\mathbf{W}_{\text{up,language}}}} \in \mathbb{R}^{C_d \times ({C_l}-{C_s})}$ represent the weights of the unique vision and language up-projection layers, while ${\mathbf{W}_{\text{up,share}}} \in \mathbb{R}^{C_d \times {C_s}}$ and ${C_s}$ are the weights and dimension of the {shared} up-projection layer. With this design, RA Adapters enable early cross-modal interactions between the two backbones.

As shown in \Cref{fig2}, freezing the pre-trained backbones and updating DARA can effectively fine-tune the baseline model with lower computation cost. Additionally, this pipeline facilitates fine-grained intra-modality representations and early inter-modality interactions.

\section{Experiments}
\label{sec:Experiments}

\subsection{Experimental Settings}
\label{sec:Experimental Settings}

\noindent \textbf{Datasets and Evaluation Metrics.} We conduct all experiments on three widely-used VG benchmarks: RefCOCO \cite{yu2016modeling}, RefCOCO+ \cite{yu2016modeling} and RefCOCOg \cite{mao2016generation}. RefCOCO has 120,624 training, 10,834 validation, and 5,657 and 5,095 images for testA and testB. RefCOCO+ consists of 120,624 training, 10,758 validation, and 5,726 and 4,889 images for testA and testB. RefCOCO+ includes some absolute-location expressions, making it more challenging. RefCOCOg has 25,779 images with 49,856 referred objects, which is divided into RefCOCOg-google (val-g) and RefCOCOg-umd (val-u and test-u). It provides longer and more complex expressions. We use Accuracy@0.5 to evaluate the performance.

\noindent \textbf{Implementation Details.} We adopt the backbone and encoder of the pre-trained DETR \cite{carion2020end} as the vision backbone, and the pre-trained BERT-base \cite{devlin2018bert} as the language backbone. The V-L Transformer is initialized with Xavier initialization. We initialize the adapters using Kaiming normal initialization, and insert them into the transformer encoder layers at the same indices as those in the DETR and BERT. The bottleneck dimensions of the DA Adapters are 128, the weight-sharing dimension $C_s$ of the RA Adapters is 256, and the adapters' scaling factor $s$ is 0.1. We fine-tune the network for 90 epochs, using the AdamW optimizer with a learning rate of $10^{-4}$ for V-L Transformer, $10^{-5}$ for the two backbones, and set the weight decay to $10^{-4}$. The learning rate will be dropped by a factor of 10 after 60 epochs. For fair comparisons, all PETL methods follow the TransVG \cite{deng2021transvg} as the baseline model, with the vision and language backbones frozen while the V-L Transformer is updated during fine-tuning.

\begin{table*}[t]
\centering

\small
\setlength{\tabcolsep}{6pt}
\setstretch{0.80}
\begin{tabular}{lcc|ccc|ccc|ccc}
\toprule
    
\multirow{2}{*}{Methods} & \multicolumn{1}{c}{Backbone} & \multicolumn{1}{c|}{\#Updated} & \multicolumn{3}{c|}{RefCOCO} & \multicolumn{3}{c|}{RefCOCO+} & \multicolumn{3}{c}{RefCOCOg} \\

 & (vision/language) & Params. & val & testA & testB & val & testA & testB & val-g & val-u & test-u \\ \midrule

\multicolumn{12}{c}{\textbf{Full fine-tuning methods}} \\ \midrule

\textbf{\textit{Two-stage:}} &  &  &  &  &  &  &  &  &  &  & \\
MAttNet \cite{yu2018mattnet} & RN101/LSTM & 47M & 76.65 & 81.14 & 69.99 & 65.33 & \textbf{71.62} & 56.02 & - & 66.58 & 67.27 \\
RvG-Tree \cite{hong2019learning} & RN101/LSTM & 47M & 75.06 & 78.61 & 69.85 & 63.51 & 67.45 & 56.66 & - & 66.95 & 66.51 \\
\midrule

\textbf{\textit{One-stage:}} &  &  &  &  &  &  &  &  &  &  & \\
FAOA \cite{yang2019fast} & DN53/LSTM & 43M & 72.54 & 74.35 & 68.50 & 56.81 & 60.23 & 49.60 & 56.12 & 61.33 & 60.26 \\
SAFF \cite{ye2021one} & DN53/BERT & 152M & 79.26 &  81.09 & 76.55 & 64.43 & 68.46 & 58.43 & - & 68.94 & 68.91 \\ 
PFOS \cite{sun2022proposal} & DN53/BERT & 152M & 77.37 & 80.43 & 72.87 & 63.74 & 68.54 & 55.84 & 61.46 & 67.08 & 66.35 \\
\midrule

\textbf{\textit{Transformer-based:}} &  &  &  &  &  &  &  &  &  &  & \\
VGTR \cite{du2022visual} & RN50/LSTM & 52M & 78.70 & 82.09 & 73.31 & 63.57 & 69.65 & 55.33 & 62.88 & 65.62 & 65.30 \\
DMRNet \cite{zhang2023one} & DN53/BERT & 152M & 76.99 & 79.71 & 72.67 & 61.58 & 66.60 & 54.00 & - & 66.03 & 66.70 \\

\rowcolor{gray!20}
TransVG$^{\dagger}$ \cite{deng2021transvg} & RN50/BERT & 151M & 80.32 & 82.67 & \textbf{78.12} & 63.50 & 68.15 & 55.63 & 66.56 & 67.66 & 67.44 \\ \midrule

\multicolumn{12}{c}{\textbf{Parameter-efficient fine-tuning methods}} \\ \midrule

Adapter \cite{houlsby2019parameter} & RN50/BERT & 3.27M & 78.02 & 79.89 & 75.23 & 61.35 & 66.34 & 54.21 & 63.18 & 65.26 & 66.65 \\
LoRA \cite{hu2022yelong} & RN50/BERT & 2.37M & 77.57 & 78.22 & 73.37 & 61.24 & 66.53 & 53.95 & 64.27 & 67.36 & 66.43 \\
AdaptFormer \cite{chen2022adaptformer} & RN50/BERT & 2.38M & 76.32 & 77.16 & 73.94 & 60.96 & 65.19 & 53.88 & 61.81 & 65.44 & 64.37 \\
CM Adapter \cite{jiang2022cross} & RN50/BERT & 3.27M & 77.37 &78.81  & 74.07 & 61.34 & 66.10 & 53.31 & 63.93 & 65.75 & 64.72 \\ 
MRS-Adapter \cite{yuan2023parameter} & RN50/BERT & 1.58M & 77.14 & 77.80 & 74.80 & 61.13 & 66.38 & 53.13 & 63.07 & 66.46 & 65.16 \\
\midrule

\rowcolor{gray!20}
DARA (ours) & RN50/BERT & 3.21M & \textbf{81.16} & \textbf{82.76} & 76.72 & \textbf{65.58} & 69.83 & \textbf{57.22} & \textbf{67.21} & \textbf{69.22} & \textbf{67.67} \\
$\Delta_{baseline}$ & RN50/BERT & \textcolor{red}{\textbf{2.13\%}} & \textcolor{red}{\textbf{+0.84}} & \textcolor{red}{\textbf{+0.09}} & \textcolor{blue}{-1.40} & \textcolor{red}{\textbf{+2.08}} & \textcolor{red}{\textbf{+1.68}} & \textcolor{red}{\textbf{+1.59}} & \textcolor{red}{\textbf{+0.65}} & \textcolor{red}{\textbf{+1.56}} & \textcolor{red}{\textbf{+0.23}} \\

\bottomrule
\end{tabular}
\vspace{-2mm}
\caption{Comparison on the RefCOCO \cite{yu2016modeling}, RefCOCO+ \cite{yu2016modeling} and RefCOCOg \cite{mao2016generation}. \#Updated Params.: the tunable parameters in the vision and language backbones. $^{\dagger}$ represents the baseline model TransVG \cite{deng2021transvg} for all parameter-efficient fine-tuning methods. $\Delta_{baseline}$ is the performance gap between our DARA and its baseline model TransVG \cite{deng2021transvg}. \textbf{Bold} denotes the best results.}
\vspace{-3mm}
\label{Table:main results}
\end{table*}

\subsection{Main Results}
\label{sec:Main Results}

The main experimental results are presented in Table \ref{Table:main results}, from which we can observe that: (1) DARA achieves the best accuracy while ensuring parameter efficiency among all methods, thus validating its effectiveness and efficiency. (2) Compared with the intra-modality efficient transferring methods such as Adapter \cite{houlsby2019parameter} and AdaptFormer \cite{chen2022adaptformer}, our method also achieves inter-modality efficient transferring, which brings significant performance improvements. (3) Compared with the most similar methods CM Adapter \cite{jiang2022cross} and MRS-Adapter \cite{yuan2023parameter}, our method considers both intra- and inter-modality knowledge transfer, thus achieving an average accuracy improvement of 4\% on the three benchmarks. (4) DARA is the only PETL method that can achieve better performance than the baseline model \cite{deng2021transvg}, by updating only \textbf{2.13\%} of the parameters in the vision and language backbones while achieving about \textbf{0.81\%} accuracy improvement. This indicates that by combining intra- and inter-modality representation transferring, the PETL method has the potential to surpass the baseline model performance to some extent. Notably, DARA does better in complex expressions, since it outperforming TransVG \cite{deng2021transvg} on all metrics in the more complex RefCOCO+ \cite{yu2016modeling} and RefCOCOg \cite{mao2016generation}, which contain more objects and more intricate language descriptions, compared to RefCOCO \cite{yu2016modeling}.

\subsection{Ablation Study and Analysis}
\label{sec:Ablation Study and Analysis}

\begin{table}[t]
\centering
\small
\setlength{\tabcolsep}{7pt}
\setstretch{0.80}
\begin{tabular}{ccc|ccc}
\toprule
\multicolumn{1}{c}{DA} & \multicolumn{1}{c}{RA} & \multicolumn{1}{c|}{Updated} & \multicolumn{3}{c}{RefCOCO}\\
Adapters & Adapters & Params. & val & testA & testB \\ \midrule

  &  & 0 & 72.72 & 73.33 & 71.27 \\
\checkmark & & 1.63M & 77.81 & 79.30 & 75.35 \\
  & \checkmark & 1.58M & 77.32 & 77.37 & 75.64 \\
\rowcolor{gray!20}
\checkmark & \checkmark & 3.21M & \textbf{81.16} & \textbf{82.76} & \textbf{76.72} \\ \bottomrule
\end{tabular}
\vspace{-2mm}
\caption{Ablations of two adapters on RefCOCO \cite{yu2016modeling} dataset.}
\vspace{-3mm}
\label{Table:ablations}
\end{table}

In this subsection, we analyze the effects of the two adapters and reasonable ways to combine and insert them into the baseline model. By default, DA Adapters are inserted sequentially after the MHA and RA Adapters are inserted in parallel with the FFN in the Transformer Encoder.

\begin{table}[t]
\centering
\small
\setlength{\tabcolsep}{5pt}
\setstretch{0.80}
\begin{tabular}{ccc|ccc}
\toprule
\multicolumn{1}{c}{Multi-head} & \multicolumn{1}{c}{Feed-forword} & \multicolumn{1}{c|}{Updated} & \multicolumn{3}{c}{RefCOCO}\\
Attention & Network & Params. & val & testA & testB \\ \midrule

DA & DA & 3.26M & 79.20 & 80.83 & 76.57 \\
RA & RA & 3.16M & 78.64 & 80.23 & 76.55 \\
RA & DA & 3.21M & 78.30 & 79.56 & 75.16 \\
\rowcolor{gray!20}
DA & RA & 3.21M & \textbf{81.16} & \textbf{82.76} & \textbf{76.72} \\ \bottomrule

\end{tabular}
\vspace{-2mm}
\caption{Comparison of different adapter combination forms on the RefCOCO \cite{yu2016modeling} dataset.}
\vspace{-3mm}
\label{Table:comparison of combination ways}
\end{table}

\noindent \textbf{Contribution of Each Component.} As shown in \Cref{Table:ablations}, freezing the two backbones leads to much greater performance degradation, compared with the PETL baselines. We also report the results of using only DA Adapters or only RA Adapters. The results indicate that adopting either of them can lead to an accuracy improvement of around 5\% compared to freezing the two backbones. By combining the DA and RA Adapters, DARA presents superior performance which demonstrates the effectiveness and efficiency of our method.

\begin{table}[t]
\centering
\small
\setlength{\tabcolsep}{10.5pt}
\setstretch{0.80}
\begin{tabular}{cc|ccc}
\toprule
\multicolumn{1}{c}{DA} & \multicolumn{1}{c|}{RA} & \multicolumn{3}{c}{RefCOCO}\\
Adapters & Adapters & val & testA & testB \\ \midrule

parallel & - & 76.32 & 77.16 & 73.94 \\
sequential & - & 77.81 & 79.30 & 75.35 \\ \midrule
sequential & sequential & 78.76 & 80.25 & 74.90 \\
\rowcolor{gray!20}
sequential & parallel & \textbf{80.26} & \textbf{82.49} & \textbf{76.65}  \\ \midrule \midrule

\multicolumn{2}{l|}{Ours ($C_s$ = 64)} & 79.00 & 80.15 & 75.96 \\ 
\multicolumn{2}{l|}{Ours ($C_s$ = 128)} & 80.26 & 82.49 & 76.65 \\ 
\rowcolor{gray!20}
\multicolumn{2}{l|}{Ours ($C_s$ = 256)} & \textbf{81.16} & \textbf{82.76} & \textbf{76.72}  \\ \bottomrule
\end{tabular}
\vspace{-2mm}
\caption{Comparison of different adapter insertion forms and the performance of RA Adapters with different weight-sharing dimensions on the RefCOCO \cite{yu2016modeling} dataset.}
\vspace{-3mm}
\label{Table:comparison of insertaion forms}
\end{table}

\noindent \textbf{Comparison of Different Combination Ways.} As shown in the first two rows of \Cref{Table:comparison of combination ways}, inserting the DA Adapters after the MHA and FFN achieves better performance than inserting the RA Adapters. This indicates that it is more important to first refine the intra-modality representations. Surprisingly, first inserting the RA Adapters in parallel with the MHA followed by the DA Adapters inserted after the FFN results in the worst performance. This demonstrates that initially sharing the adapters’ weights may impair the rich pre-trained intra-modality representations. Consequently, forcibly transferring these impaired representations to the VG domain leads to worth performance. In contrast, DARA first transfers the intra-modality representations to be more fine-grained for VG, and then shares the adapters' weights to bridge the two modalities, thereby achieving the best performance.

\noindent \textbf{Comparison of Different Insertion Forms.} Motivated by AdaptFormer \cite{chen2022adaptformer}, we investigate \textbf{parallel} (inserted in parallel) versus \textbf{sequential} (inserted sequentially) insertion formulations. As the upper part of \Cref{Table:comparison of insertaion forms} shows, sequential DA Adapters outperform parallel design by effectively transferring the pre-trained intra-modality representations from the MHA for VG. Meanwhile, the parallel design of RA Adapters efficiently bridges the two backbones to capture the inter-modality representations, resulting in better performance. Furthermore, we explore the impacts of varying degrees of cross-modal interactions, shown in the lower part of \Cref{Table:comparison of insertaion forms}. As CM Adapter \cite{jiang2022cross}  shows the best text-video retrieval performance by sharing half the visual dimensions, we find fully sharing the 256-dim visual features achieves the highest VG accuracy. We hypothesize that VG serves as a more challenging vision-language task, therefore, requiring much deeper cross-modal interactions.

\begin{figure}[t]
\centering
\includegraphics[width=0.48\textwidth]{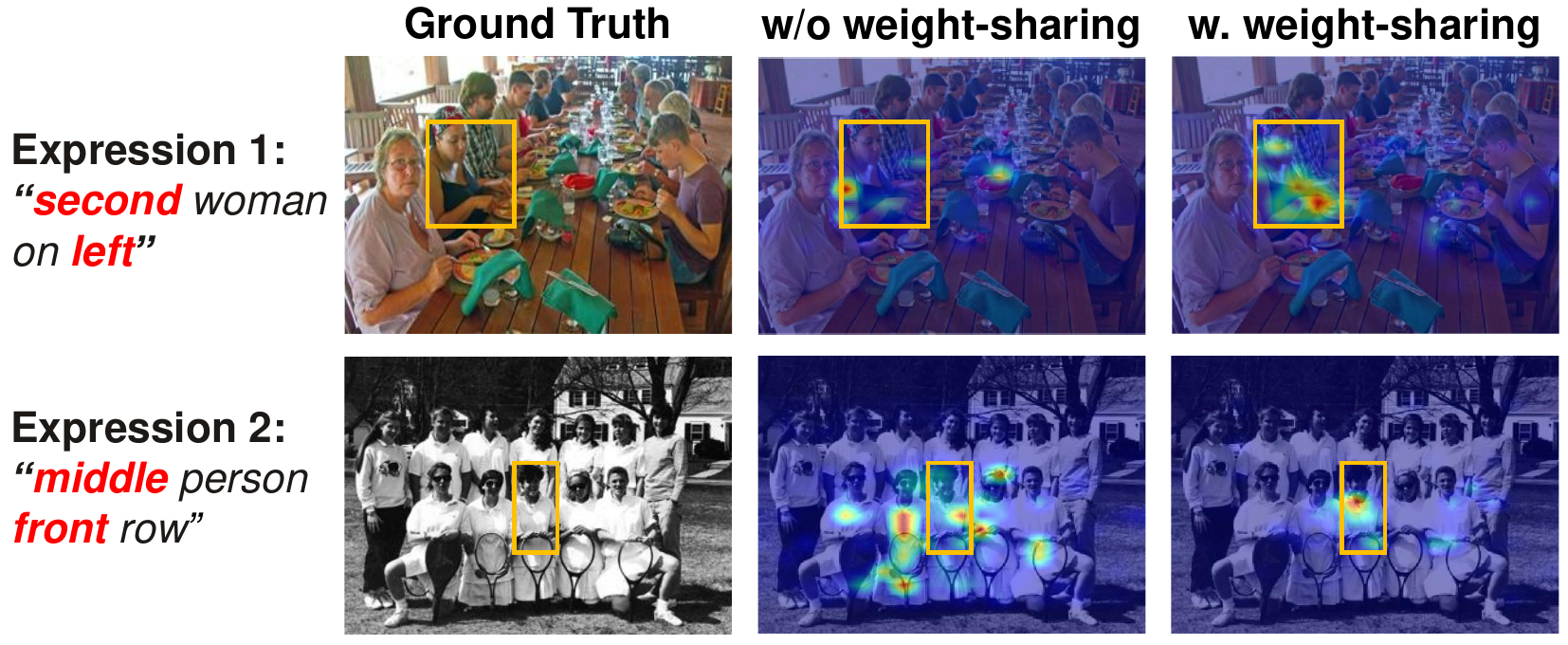}
\vspace{-7mm}
\caption{Visualizations of cross-attention maps between the \texttt{[REG]} token and visual tokens under different strategies.}
\vspace{-3mm}
\label{fig3}
\end{figure}

\noindent \textbf{Effectiveness of Weight-sharing Strategy.} To further study the impact of early cross-modal interactions via RA Adapters, we compare DARA with a baseline without weight-sharing on challenging cases requiring spatial reasoning. We visualize the cross-attention maps between the \texttt{[REG]} token and visual tokens in the last layer of the V-L Transformer. As depicted in \Cref{fig3}, DARA's \texttt{[REG]} token focuses more on the referred object than the baseline. This suggests that weight-sharing strategy of RA Adapters promotes the synergy between vision and language modalities, thereby enhancing spatial reasoning for visual grounding.

\section{Conclusion}
\label{sec:conclusion}
In this paper, we present a new parameter-efficient tuning method DARA for visual grounding. DARA first utilizes Domain-aware Adapters to enhance the pre-trained intra-modality representations, making them fine-grained for the domain of VG. Sequentially, it employs Relation-aware Adapters to bridge the relation between the vision and language backbones and capture inter-modality representations. In this way, DARA achieves the best accuracy while saving numerous updated parameters across three benchmarks, and even surpasses the performance of the baseline model.

\section{Acknowledgments}
This research was partially supported by the National Natural Science Fund of China (Grant Nos. 62306329 and 62103425), Natural Science Fund of Hunan Province (Grant Nos. 2023JJ40676 and 2022JJ40559), and Sichuan Science and Technology Program (No. 24QYCX0399).


{\small
\bibliographystyle{IEEEbib}
\bibliography{refs}
}
\end{document}